\documentclass[conference]{IEEEtran}
\usepackage{cite}
\usepackage{graphicx,epstopdf,url}
\usepackage{subfigure}
\usepackage{bbold}
\usepackage{verbatim}
\usepackage{textcomp}
\usepackage{slashbox}
\usepackage{amsmath}
\usepackage{amssymb}
\usepackage{multirow}
\usepackage{booktabs}
\usepackage[utf8]{inputenc}  
\usepackage[lined,ruled,boxed,commentsnumbered,linesnumbered]{algorithm2e}
\usepackage{indentfirst}
\usepackage{bbm}
\usepackage[all]{xy}
\usepackage{amsmath}

\usepackage{array}
\newcolumntype{L}[1]{>{\raggedright\let\newline\\\arraybackslash\hspace{0pt}}m{#1}}
\newcolumntype{C}[1]{>{\centering\let\newline\\\arraybackslash\hspace{0pt}}m{#1}}
\newcolumntype{R}[1]{>{\raggedleft\let\newline\\\arraybackslash\hspace{0pt}}m{#1}}

\begin{document}
\title{Calibration-Aided Edge Inference Offloading via Adaptive Model Partitioning\\ of Deep Neural Networks}

\author{
    \IEEEauthorblockN{Roberto G. Pacheco\IEEEauthorrefmark{1}, Rodrigo S. Couto\IEEEauthorrefmark{1}~\IEEEmembership{Member,~IEEE}, and Osvaldo~Simeone\IEEEauthorrefmark{2}~\IEEEmembership{Fellow,~IEEE}}
   \IEEEauthorblockA{\IEEEauthorrefmark{1}Universidade Federal do Rio de Janeiro, PEE/COPPE/GTA, Rio de Janeiro, RJ, Brazil\\
    Email: pacheco@gta.ufrj.br, rodrigo@gta.ufrj.br}
   \IEEEauthorblockA{\IEEEauthorrefmark{2}King's College London, KCLIP lab, Department of Engineering, London, United Kingdom\\
    Email: osvaldo.simeone@kcl.ac.uk}
}

\maketitle

%\IEEEpeerreviewmaketitle

\begin{abstract}
Mobile devices can offload deep neural network (DNN)-based inference to the cloud, overcoming local hardware and energy limitations. However, offloading adds communication delay, thus increasing the overall inference time, and hence it should be used only when needed. An approach to address this problem consists of the use of adaptive model partitioning based on early-exit DNNs. Accordingly, the inference starts at the mobile device, and an intermediate layer estimates the accuracy: If the estimated accuracy is sufficient, the device takes the inference decision; Otherwise, the remaining layers of the DNN run at the cloud. Thus, the device offloads the inference to the cloud only if it cannot classify a sample with high confidence. This offloading requires a correct accuracy prediction at the device. Nevertheless, DNNs are typically miscalibrated, providing overconfident decisions. This work shows that the employment of a miscalibrated early-exit DNN for offloading via model partitioning can significantly decrease inference accuracy. In contrast, we argue that implementing a calibration algorithm prior to deployment can solve this problem, allowing for more reliable offloading decisions.
\footnote{\textcopyright2021 IEEE. Personal use of this material is permitted. Permission from IEEE must obtained for all other uses, in any current or future media, including reprinting/republishing this material for advertising or promotional purposes, creating new collective works, for resale or redistribution to servers or lists, or reuse of any copyrighted component of this work in other works.} 
\end{abstract}

\section{Introduction}
\label{sec:intro}

Deep neural networks (DNNs) are becoming essential tools to carry out sophisticated inference tasks on data. These include computer vision applications enabling smart vehicles that can use data gathered by cameras and sensors to detect objects and pedestrians on the road, avoiding accidents~\cite{bechtel2018deeppicar}.

Mobile devices generally offload DNN inference to a cloud computing infrastructure. The cloud can provide the required computational resources to accelerate the inference, such as Graphics Processing Units (GPUs)~\cite{satyanarayanan2017emergence,kang2017neurosurgeon}. In cloud-based scenarios, end devices gather raw data and send it to the cloud, which executes the inference. However, cloud offloading is highly dependent on network conditions. It may result in excessive communication delays between end devices and the cloud server due to network congestion and communication channel degradation.  

Edge computing can reduce the communication delay imposed by cloud offloading by directly using the computational resources available on the end device or close to it, such as at a base station~\cite{satyanarayanan2017emergence}. For example, edge computing can leverage the processing power available on smartphones and wearables to execute DNN inference locally~\cite{xu2019deepwear}. Nevertheless, the processing power of end devices is significantly lower as compared to the cloud server. Thus, when inference requires complex computations, such as detecting scenes involving many objects, edge computing may yield insufficient accuracy. 

\begin{figure}[!ht]
\center
\includegraphics[width=\linewidth]{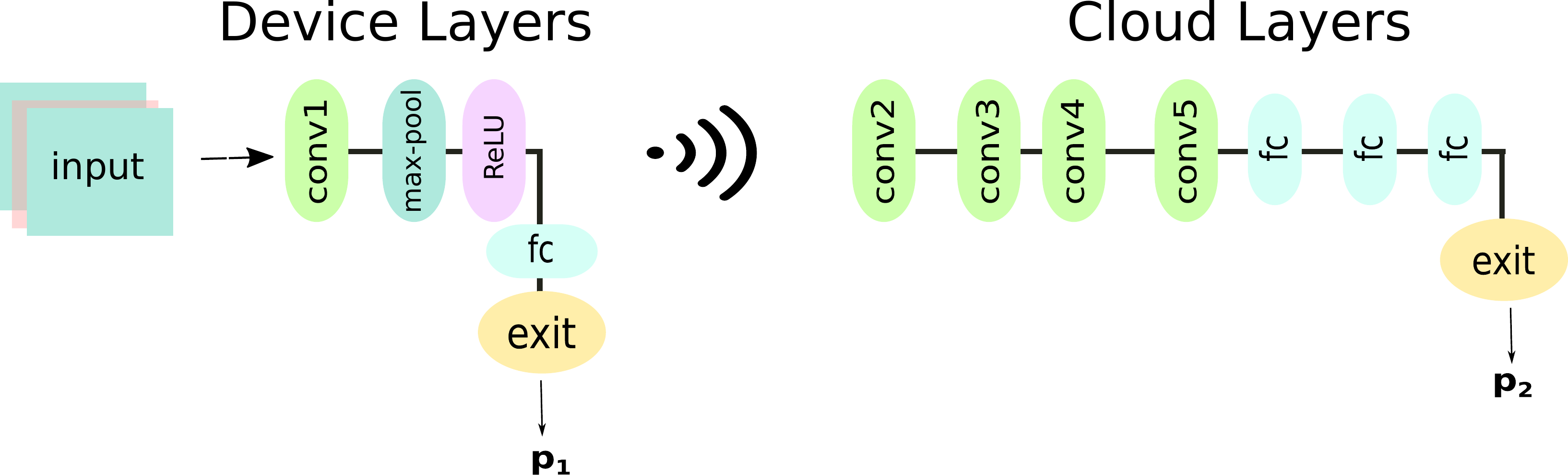}
\caption{Illustration of a B-AlexNet with an early-exit branch and offloading to the cloud via model partitioning.}
\label{fig:fixed_partitioning}
\end{figure}

In a standard DNN, inference needs to process the input data through all the DNN's layers. Different works propose DNNs with ``early exits'', enabling decision making after a limited number of layers on less complex inputs. These include BranchyNet~\cite{teerapittayanon2016branchynet} and SPINN~\cite{laskaridis2020spinn}. In these works, early-exit points classify a sample at an intermediate layer when the estimated inference accuracy is larger than a tunable threshold. Therefore, an inference task can be adaptively completed earlier on simpler inputs, saving further computations. 

A related idea that has found an application for cloud offloading is DNN model partitioning~\cite{kang2017neurosurgeon, hu2019dynamic, pacheco2020inference}. DNN model partitioning uses optimization to determine a layer, known as the partitioning layer, that splits a DNN into two parts: The edge device process the layers before the partitioning layer. In contrast, the cloud process the last layers. The partitioning layer choice aims a given objective, such as reducing the inference time or saving energy on the edge device.

In this paper, we address the need to ensure high accuracy while reducing inference latency by combining early-exit DNNs with model partitioning~\cite{pacheco2020inference,laskaridis2020spinn}. This solution enables adaptive offloading as a function of the current input's complexity: If early-exit points in the DNNs implemented at the device cannot provide sufficient accuracy, the data is sent to the cloud for processing the remaining layers. 

Adaptive model partitioning via early-exit DNNs requires that the system provides an accurate measure of uncertainty. This measure verifies if the inference decision is confident enough to be completed on an early exit~\cite{kendall2017uncertainties}. Consequently, early-exit DNNs should provide the necessary quantification of accuracy via the probabilities computed by a softmax layer. However, 
Guo~\textit{et al.}~\cite{guo2017calibration} have shown that modern DNNs are not well-calibrated, providing overconfident decisions. Note that Guo~\textit{et al.} do not address or analyze the calibration problem using DNNs with early exits. In this paper, we address the problem that a miscalibrated early-exit DNN can erroneously decide to classify a sample at the edge, decreasing accuracy and reliability.

Specifically, in this work, we address miscalibration for adaptive model partitioning via early-exit DNNs. We demonstrate that miscalibration can significantly affect the reliability of offloading decisions. Then, we apply a state-of-the-art calibration method suggested by Guo~\textit{et al.}~\cite{guo2017calibration} on an early-exit DNN. We evaluate its impact on offloading reliability, demonstrating its effectiveness in ensuring that the device offloads only samples that need additional computation to achieve the desired accuracy level. We note that reference~\cite{laskaridis2020spinn} applied calibration for offloading. However, it does not provide any discussion and illustration of the actual impact of this step on the reliability and effectiveness of mobile offloading. This analysis is the focus of our paper.

This paper is structured as follows. We review related work in Section~\ref{sec:related_work}. Then, Section~\ref{sec:mobile_edge_inference} describes the edge inference scenario under study. Section~\ref{sec:experiments} analyzes the impact of miscalibration and calibration prior to deployment. Finally, Section~\ref{sec:conclusion} concludes this work and presents the next steps.

\section{Related Work}
\label{sec:related_work}

Over the years, DNNs have presented significant improvements in terms of accuracy. However, Guo~\textit{et al.}~\cite{guo2017calibration} show that this accuracy improvement often results in miscalibrated decisions. Focusing on classification, the confidence level of a DNN on a given sample is its estimated probability of correct inference evaluated at the last, softmax, layer. A DNN is miscalibrated if the inferred class's probability does not reflect its ground-truth posterior probability for a given input. A perfectly calibrated DNN is one for which confidence level corresponds precisely to the posterior distribution, that is, to the classification accuracy. For example, under perfect calibration, a DNN that outputs a confidence value of 0.8 for a subset of inputs would give the correct decision for 80\% of these samples. To solve the calibration problem, Guo~\textit{et al.} evaluate different calibration methods over several datasets. Their work concludes that the \emph{Temperature Scaling} method is often the most effective for obtaining well-calibrated probabilities for computer vision applications.  Guo~\textit{et al.}~\cite{guo2017calibration} do not address the calibration problem of early-exit DNNs. This analysis is vital for applications that involve edge inference offloading.

Early-exit DNN is a growing research topic, whose goal is to accelerate inference time by reducing processing delay. The idea is to insert ``early exits'' in a DNN architecture, classifying samples earlier at its intermediate layers if a sufficiently accurate decision is predicted. To this end, an early-exit layer generates a probability vector, such as the exit layer of a traditional DNN. The classification confidence is estimated from this probability vector to decide if a sample is confident enough to be classified at the corresponding intermediate layer. Therefore, samples easier to classify can exit earlier, saving further computations. BranchyNet~\cite{teerapittayanon2016branchynet} is an example of an early-exit DNN which uses an uncertainty metric, such as entropy, to decide whether a sample can be classified earlier. BranchyNet verifies if the entropy value is less than a threshold. If so, BranchyNet inference stops, and the sample is classified. This work shows that these early exits can classify a large portion of the input samples.

SPINN~\cite{laskaridis2020spinn} is a DNN partition system which considers the employment of early-exit DNNs.  SPINN~\cite{laskaridis2020spinn} uses the classification confidence, i.e., the maximum of the softmax layer, as the confidence criterion. As we discussed, in a perfectly calibrated network, this criterion matches exactly the accuracy~\cite{guo2017calibration}. Early-exit DNNs are also considered in \cite{wang2019dynexit}, which uses a residual neural network (ResNet) implemented on specific hardware. Other works propose an optimization problem to select the most suitable DNN depth for model partitioning. For example, Edgent~\cite{li2019edge} reduces the inference time, while the optimization in~\cite{stamoulis2018designing} saves energy. 

None of those above works analyses the calibration problem in early-exit DNNs. For example, SPINN has a calibration step. However, they do not evaluate the actual impact of this step. Similar to SPINN~\cite{laskaridis2020spinn}, we consider the partitioning of early-exit DNNs. We complement this work by analyzing the impact of miscalibration and evaluating the potential improvements obtained with a calibration method.    

\section{Edge Inference Offloading}
\label{sec:mobile_edge_inference}

DNNs with early exits have side branches that allow inference to be terminated at intermediate layers. The classification accuracy is thus estimated to decide if a given side branch can correctly predict each sample. As we discussed, the early classification can reduce the processing delay and, thus, the inference time. 

Once trained, DNNs with early exits can receive an input, say an image, to classify. The input is processed, layer-by-layer, until it reaches a $i$-th side branch. Then, on the $i$-th side branch, a fully-connected layer generates the logit vector $\boldsymbol{z}_{i}$, that is used to obtain the probability vector $\boldsymbol{p}_{i}$ through the softmax layer, as in
\begin{equation}
\label{eq:probability_vector}
\boldsymbol{p}_{i} = \text{softmax}(\boldsymbol{z}_{i}) \propto \exp(\boldsymbol{z}_{i}),
\end{equation} where the exponential function is applied element-wise. The probability vector $\boldsymbol{p}_{i}$ collects the probabilities that a sample belongs to any of the predefined classes. The class with the largest probability in vector $\boldsymbol{p}_{i}$ corresponds to the inferred class, and its confidence level is given by the corresponding probability $\max \boldsymbol{p}_{i}$. 

For each sample, the device at the edge verifies if the confidence level $\max \boldsymbol{p}_{i}$ is greater than a predefined target confidence $p_{\text{tar}}$. If so, the side branch can classify the sample, and the inference stops. Consequently, the input sample is no longer processed by the next layers, reducing the processing delay and avoiding offloading. Otherwise, if the confidence value is less than $p_{\text{tar}}$, the sample is processed by the next layers until it reaches the next side branch and follows the same procedure as described before. If no side branches reach $p_{\text{tar}}$, the cloud executes the last DNN layers. 

In this work, we employ a B-AlexNet architecture, which consists of an AlexNet~\cite{krizhevsky2012imagenet}, trained using the BranchyNet methodology~\cite{teerapittayanon2016branchynet} for early-exit DNNs. The training uses the CIFAR-10 dataset~\cite{cifar10}, consisting of 32$\times$32 color images divided into ten classes. In our analysis, we use 45,000 training images, 3,000 validation images, and 7,000 test images. We employ the Pytorch\footnote{https://pytorch.org/} framework in our analysis. 

Figure~\ref{fig:fixed_partitioning} illustrates the system under study. Unless stated otherwise, we will consider a DNN having one side branch, located after the first ReLU layer. The figure does not include all the layers for better clarity. The vector $\boldsymbol{p}_{1}$ refers to the probability vector generated by the side branch at the mobile device, which is also the last layer implemented at the edge. In the same way, $\boldsymbol{p}_{2}$ is the probability vector generated by the main exit, implemented at the cloud. Accordingly, the mobile device processes the first convolutional layer, while the last ones execute at the cloud.

\section{DNN calibration Analysis}
\label{sec:experiments}

The analysis presented in this section evaluates the impact of calibrating a DNN with early exits, according to the desired confidence target $p_{\text{tar}}$. To this end, we apply the Temperature Scaling method to calibrate the side branch.

\subsection{Temperature Scaling}
\label{sec:temperature_scaling}

Temperature Scaling is a post-processing calibration method whose goal is to generate a calibrated probability vector $\hat{\boldsymbol{p}}_{i}$ for a side branch~\cite{guo2017calibration}. Temperature Scaling uses a single scalar parameter $T$ to obtain the calibrated vector\begin{equation}
\hat{\boldsymbol{p}}_{i}=\text{softmax}\big(\frac{\boldsymbol{z_i}}{T}\big). 
\end{equation} The parameter $T$ is determined using the validation dataset to minimize the negative log-likelihood on the validation dataset, for fixed weights. We apply this approach to the side branch at the edge devices. Next, we present our calibration analysis, which uses samples from the test dataset.    

\subsection{Offloading Probability}
\label{subsec:calibration_exited_sample}

\begin{figure}[!ht]
\center
\includegraphics[width=0.75\linewidth]{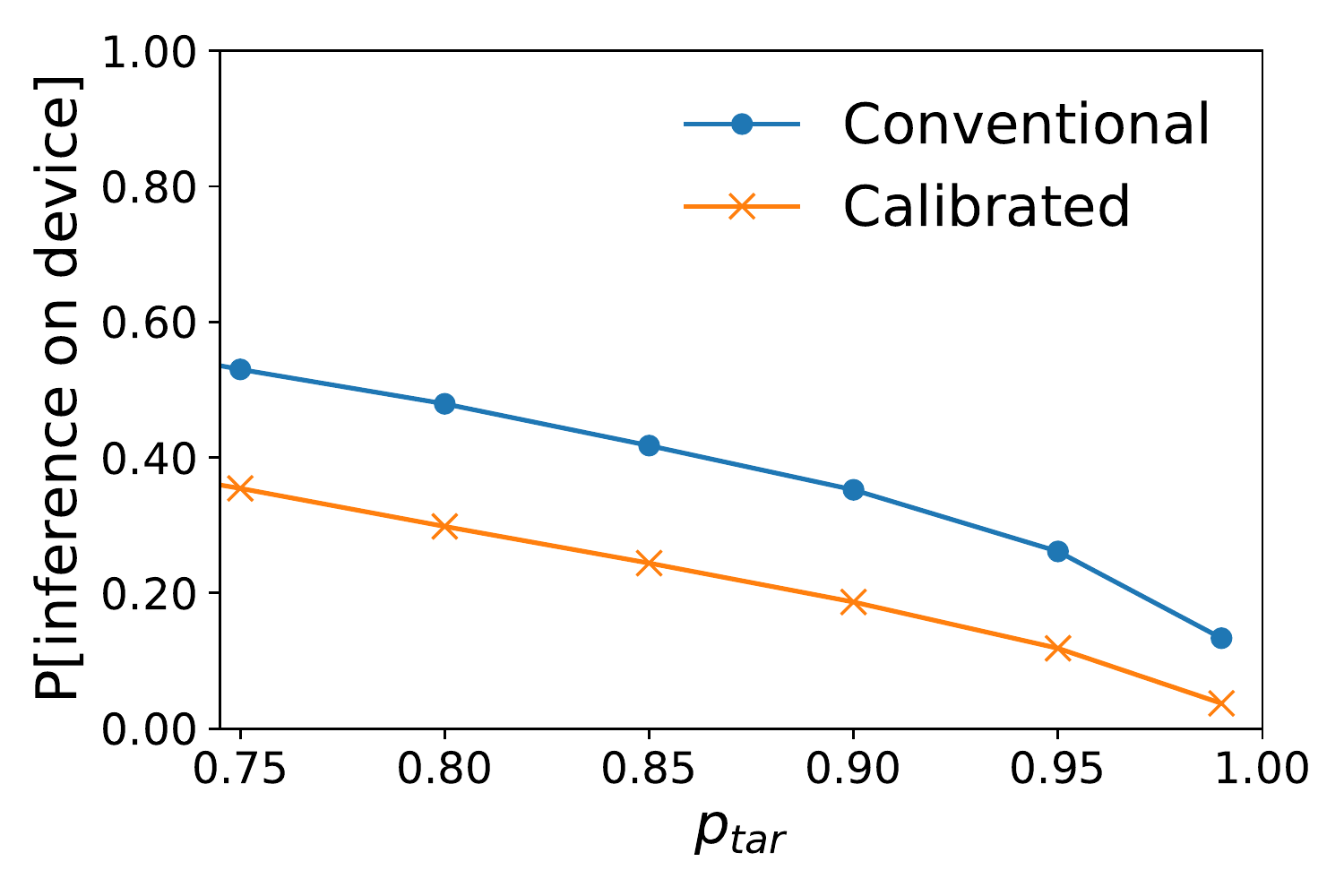}
\caption{Probability of classifying samples on the device, i.e., the complement of the offloading probability, using a conventional DNN.}
\label{fig:prob_inference_on_device}
\end{figure}

We start by evaluating the impact of calibration on the offloading probability. For a given desired target accuracy $p_{\text{tar}}$,  we evaluate the number of samples classified on the device divided by the total number of samples in the test dataset. Figure~\ref{fig:prob_inference_on_device} shows the probability of classifying on the device -- the complement of the offloading probability -- as a function of $p_{\text{tar}}$, using a conventionally trained network and a calibrated side branch. Calibration reduces the probability of classifying samples on the device as compared to a conventional DNN. As we will see, this is because a conventionally trained DNN overestimates the reliability, classifying more samples than it should, to ensure the accuracy level $p_\text{tar}$. 

\subsection{Accuracy Before and After Calibration}
\label{subsec:accuracy}

\begin{figure}[!ht]
\centering
\subfigure[Accuracy at the device versus confidence.]{\label{fig:accuracy_device_vs_confidence}\includegraphics[width=0.75\linewidth]{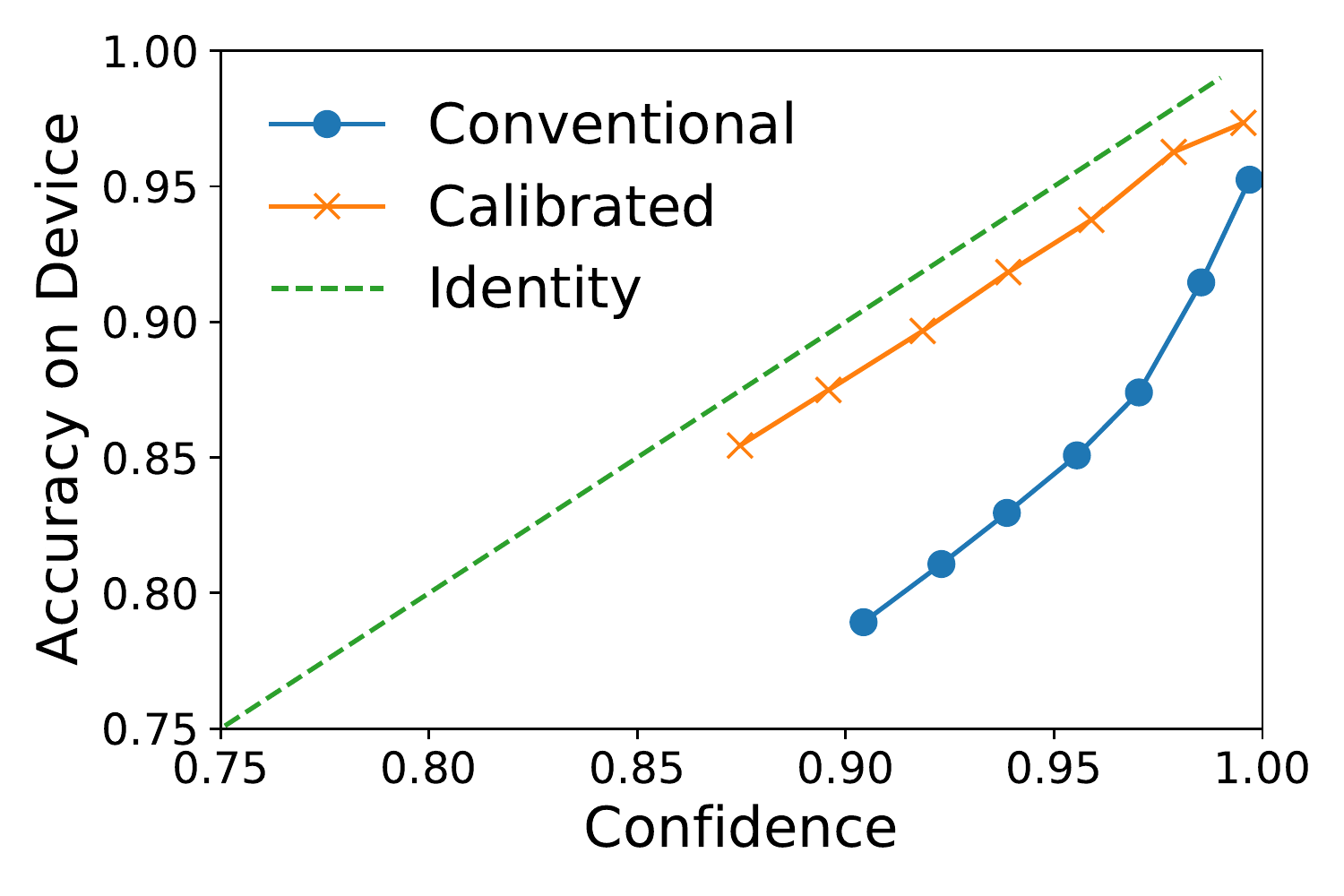}}
\subfigure[Accuracy at the device.]{\label{fig:accuracy_branches1}\includegraphics[width=0.75\linewidth]{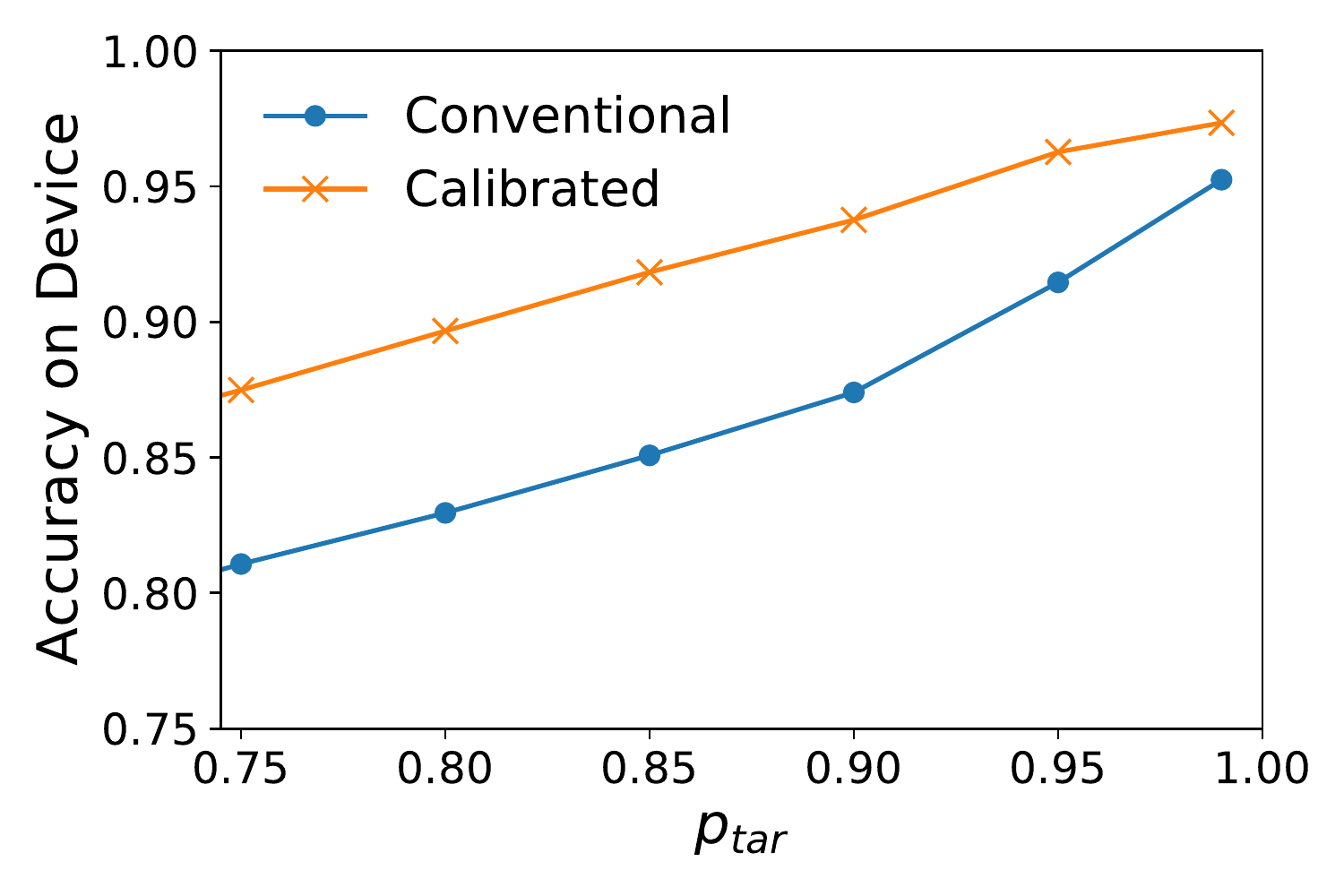}}
\subfigure[Total accuracy.]{\label{fig:accuracy_total}\includegraphics[width=0.75\linewidth]{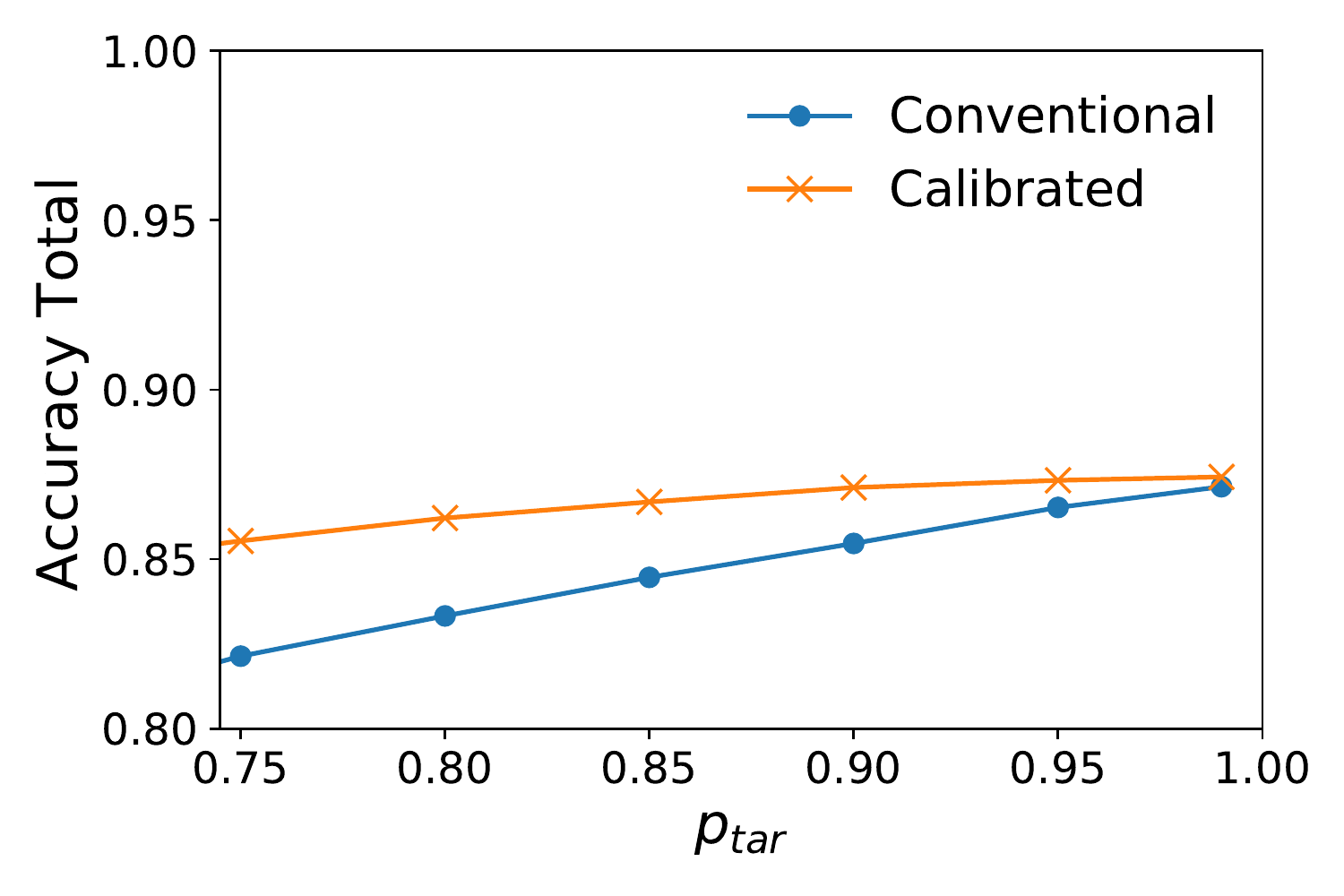}}
\caption{Confidence, accuracy at the device, and overall accuracy for conventional and calibrated DNNs.}
\label{fig:accuracy_confidence_before_after_calibration}
\end{figure}

First, Figure~\ref{fig:accuracy_device_vs_confidence} presents the accuracy on device as a function of the confidence level. We obtain each point in this figure for a given $p_\text{tar}$ configuration, under conventional and calibrated training. For reference, we also show the ``identity'' curve, in which the accuracy is precisely equal to the average confidence. This curve represents the case of ideal calibration when the confidence provided by the side branch reflects exactly the ground-truth correctness. The figure shows that calibration can better approximate the confidence from the accuracy. Thus, a calibrated side branch can provide confidence that better reflects the ground-truth accuracy. 

The impact of calibration on the accuracy obtained at the device as a function of $p_\text{tar}$ is shown in Figure~\ref{fig:accuracy_branches1}.
This result shows that, for all values of $p_{\text{tar}}$, calibration achieves a more considerable accuracy than conventional training. As discussed, this is because conventional training overestimates its confidence, yielding lower accuracy than required. For the calibrated network, the accuracy is always greater or close to $p_\text{tar}$. These results indicate that, under calibration, the tunable parameter $p_{\text{tar}}$ works as a reliable target for accuracy on the device, whereas this is not the case for conventional training. 

Finally, Figure~\ref{fig:accuracy_total} shows the accuracy across all samples, including those classified at the cloud. This result shows that a calibrated DNN can outperform a conventional DNN, for any $p_{\text{tar}}$ configuration.

\subsection{Inference Outage Probability}
\label{subsec:probability_outage}

As seen in the previous experiment, the tunable parameter $p_{\text{tar}}$ serves as a reliable estimate of the accuracy obtained with an early exit at the device. To further analyze the impact of calibration on the accuracy requirement, we introduce here the notion of inference outage. We define an outage as an event that occurs when the device's accuracy is smaller than the target $p_{\text{tar}}$. The outage probability measures the reliability of offloading decisions.

\begin{figure}[!ht]
\centering
\includegraphics[width=0.75\linewidth]{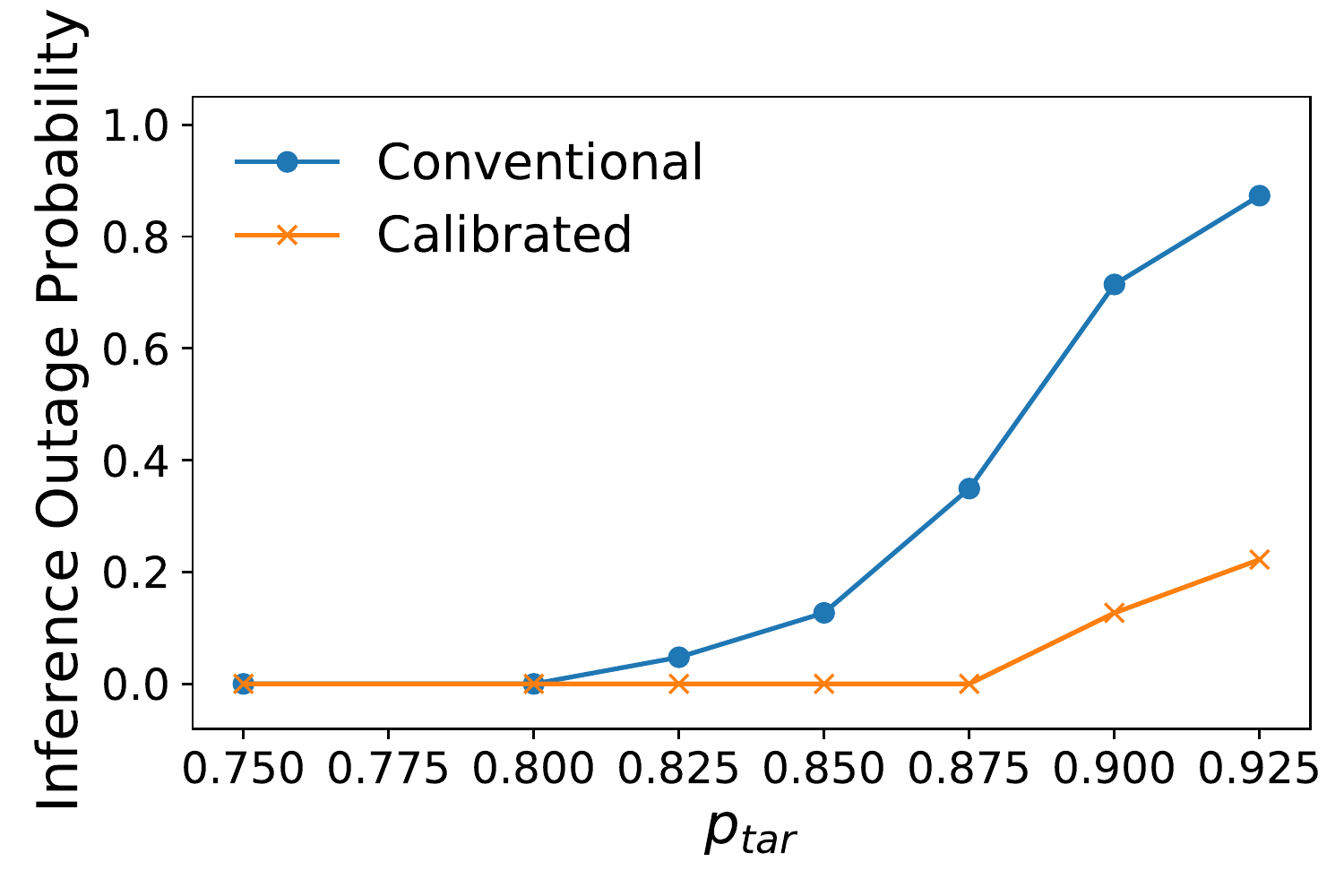}
\caption{Inference outage probability on the device, using conventional and calibrated DNNs.}
\label{fig:outage_before_after_calibration_branch_1}
\end{figure}

To calculate the inference outage, we divide the test dataset again into batches with 512 images each. For each batch, we evaluate the average accuracy for the samples classified at the device. Figure~\ref{fig:outage_before_after_calibration_branch_1} shows the inference outage probability using conventional and a calibrated DNN versus $p_{\text{tar}}$. We can observe that for all values of the target probability $p_{\text{tar}}$, calibration improves the outage probability. Hence, a calibrated side branch is more likely to meet the reliability target $p_{\text{tar}}$ than a conventional one. Moreover, we note that the outage probability increases with $p_{\text{tar}}$ for both conventional and calibrated branches. However, the outage probability using a calibrated branch remains zero for any $p_{\text{tar}}$ less than 0.875. For the same $p_{\text{tar}}$ value, the outage probability is close to 0.4 for a conventional branch. 

\subsection{Missed Deadline Probability}
\label{subsec:missed_deadline}

Offloading should not only be reliable, but it should also meet latency constraints. To evaluate the calibration impact on latency, we introduce here the missed deadline probability. We say that a missed deadline event occurs when the inference time, defined as the overall time required to infer a batch of samples, is larger than an application-defined deadline $t_{\text{tar}}$ or when the total batch accuracy is smaller than $p_{\text{tar}}$. The rationale for this definition is that inference fulfills the application's requirements only if both latency target $t_{\text{tar}}$ and reliability target $p_{\text{tar}}$ are met for each batch. Note that, when conditioned on the samples classified at the device, the missed deadline probability is by definition no smaller than the inference outage probability. In this section, however, we focus on the overall end-to-end application performance and consider all samples classified at the device and the cloud. 

To calculate the missed deadline probability, we divide the test dataset again into batches with 512 images. Then, we measure the overall inference time as the time required to process all the samples in one batch, comprising both computing and communication delays. For the contribution of computing to the latency, we estimate the processing delay on the edge device based on the values reported in~\cite{colburn2019optical}. This paper measures the processing delay required by each layer of an AlexNet using an Intel i7 CPU. For the computing time of cloud processing, the analysis uses the computational resources provided by Google Colaboratory\footnote{\url{https://colab.research.google.com/}}. Specifically, for this experiment, we use an Intel 2-core Xeon(R)@ 2.20GHz processor, 12 GB VRAM, and a GPU NVIDIA Tesla K80, provided by the Google platform. We also account for the communication delay, which is accrued when sending the partitioning layer's output data from the device to the cloud. The communication delay is given by the output data size divided by the average uplink rate. We evaluate the data size as the size of data structures in Pytorch, and we use the average uplink rate of 18.8\,Mbps, derived from a Wi-Fi scenario in~\cite{hu2019dynamic}. 

For each sample in a batch, if classification occurs at the device, only processing delay is accounted for, while classification at the cloud entails processing latency at the edge device, communication latency, and processing latency at the cloud. If the average inference time is larger than the application-defined deadline $t_{\text{tar}}$, or the batch accuracy is smaller than $p_{\text{tar}}$, a missed deadline event occurs. The missed deadline probability is obtained by averaging over all batches. 

\begin{figure}[!ht]
\centering
\subfigure[$p_{\text{tar}}=0.75$.  ]{\label{fig:missed_deadline_branch_75}\includegraphics[width=0.75\linewidth]{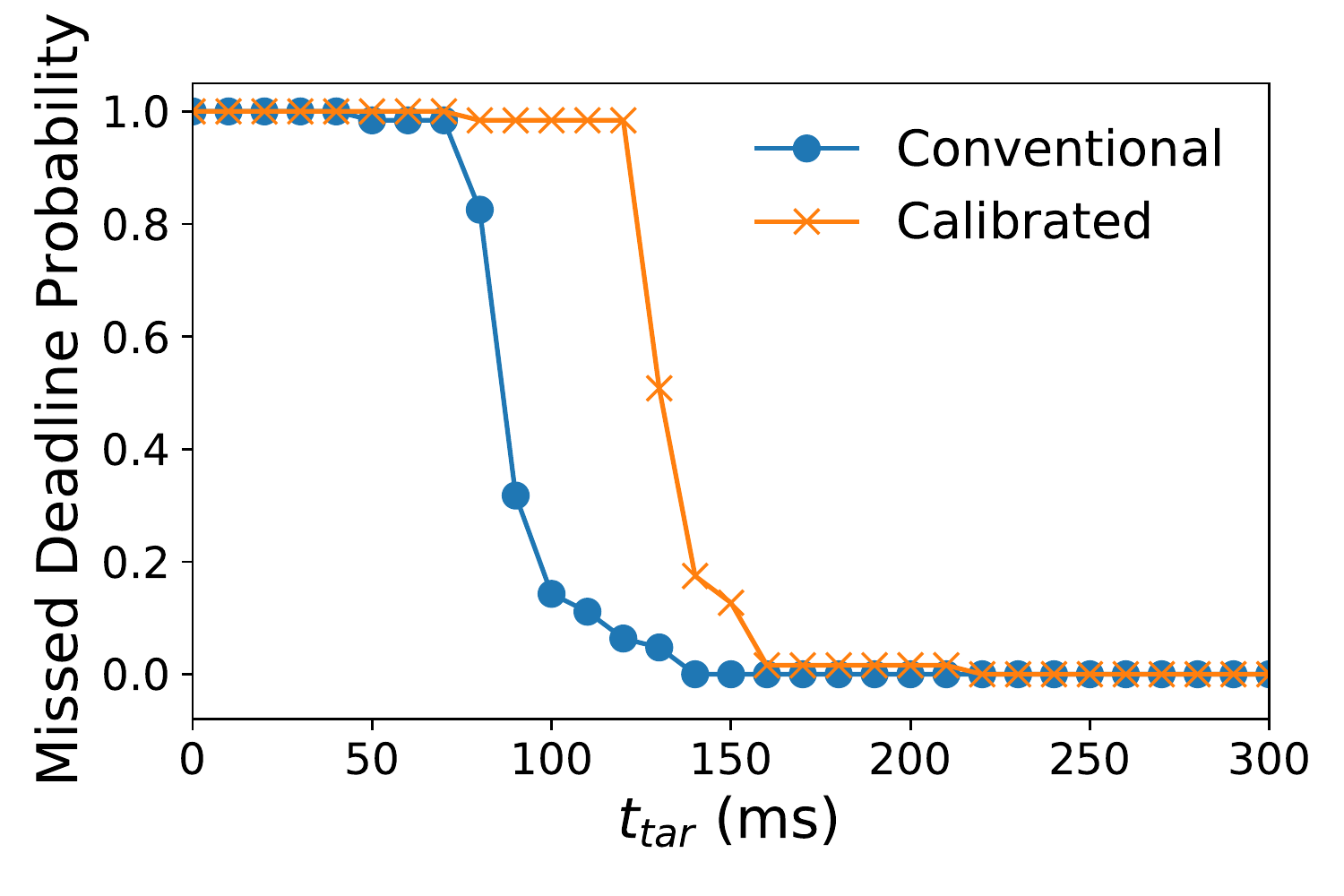}}
\subfigure[$p_{\text{tar}}=0.825$.]{\label{fig:missed_deadline_branch_825}\includegraphics[width=0.75\linewidth]{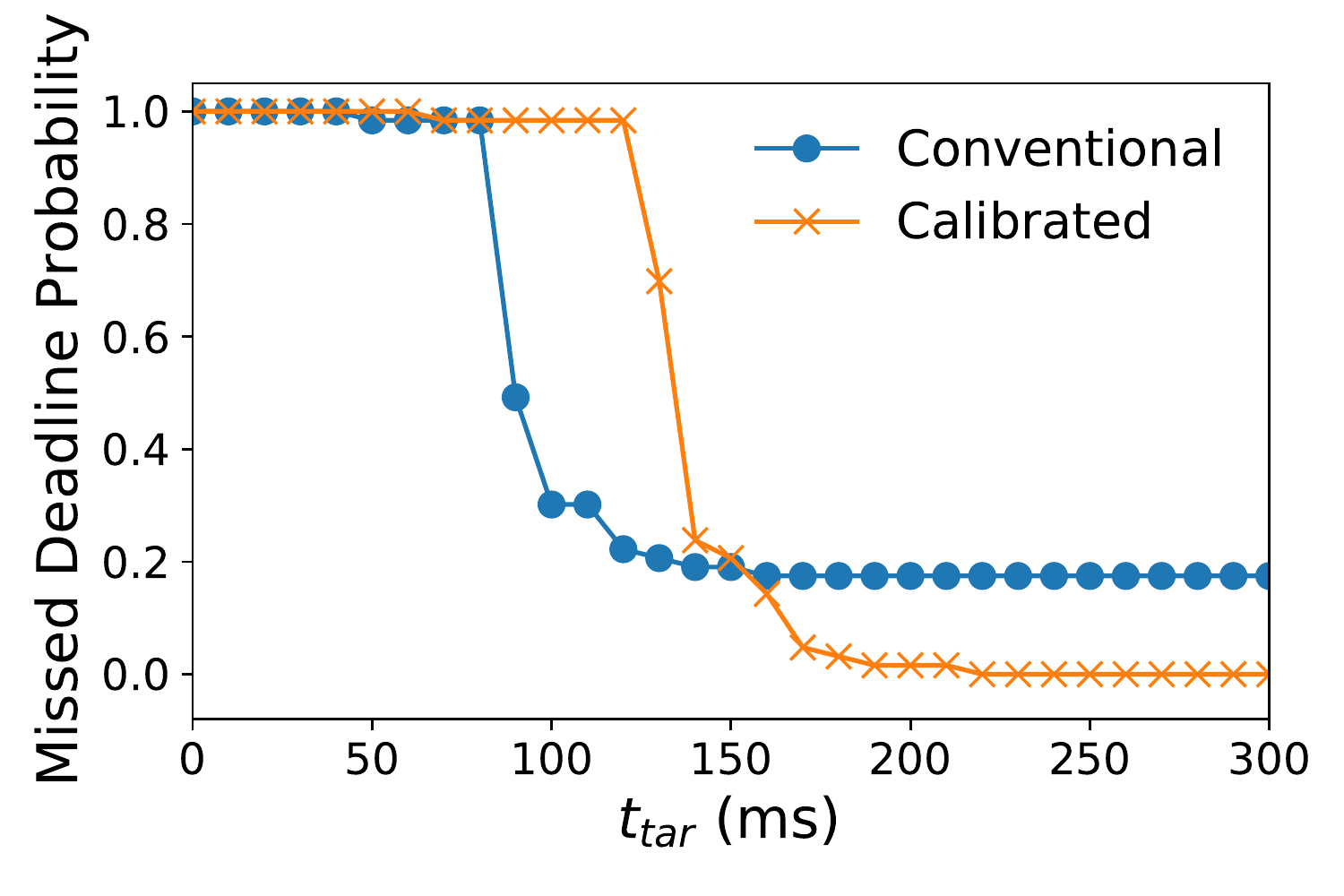}}
\subfigure[$p_{\text{tar}}=0.85$.]{\label{fig:missed_deadline_branch_85}\includegraphics[width=0.75\linewidth]{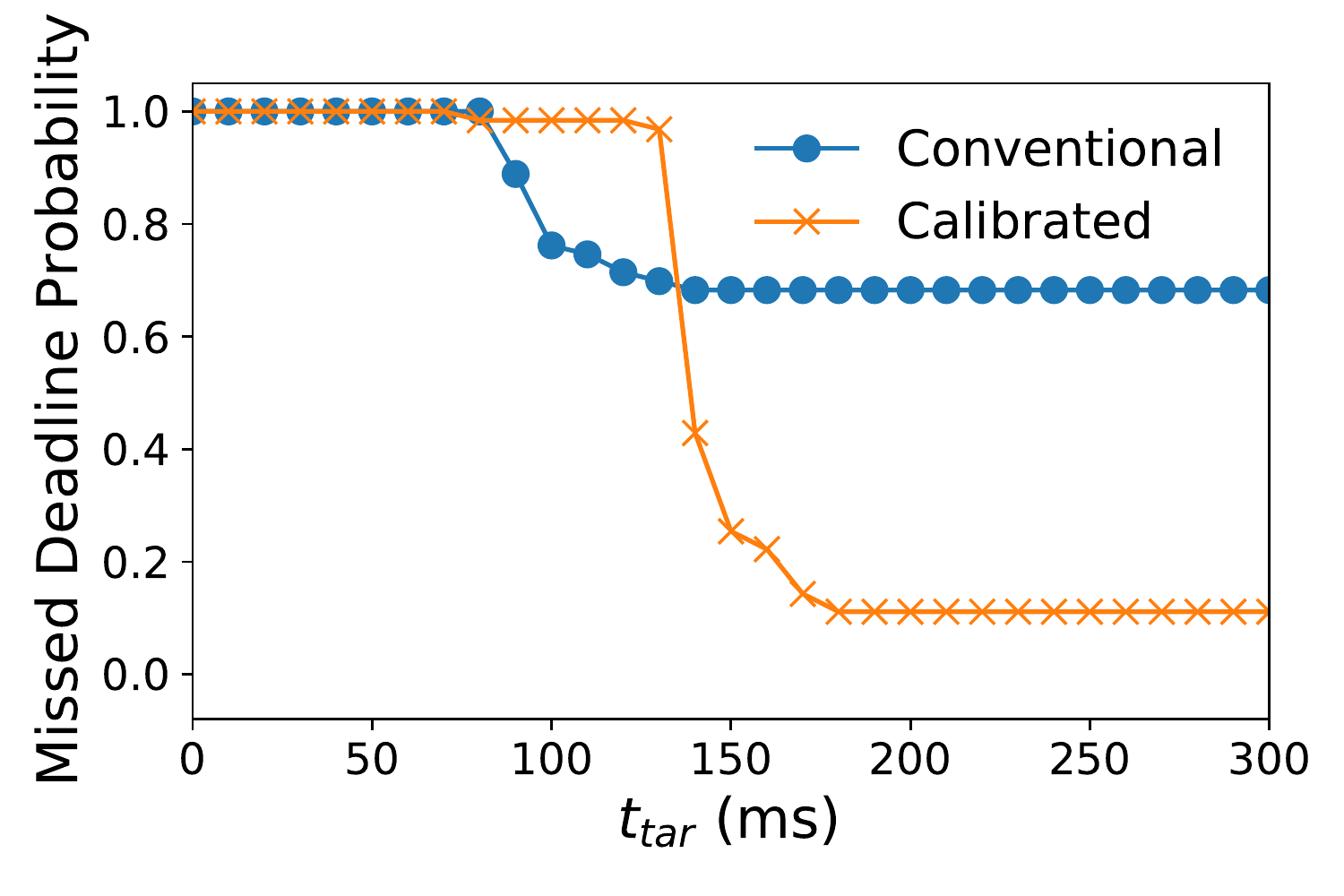}}
\caption{Missed deadline probability for a conventional and a calibrated DNN.}
\label{fig:missed_deadline_calibration}
\end{figure}

Figure~\ref{fig:missed_deadline_calibration} shows the missed deadline probability as a function of $t_{\text{tar}}$ for different requirements on $p_{\text{tar}}$. The results for $p_{\text{tar}} = 0.75$ in Figure~\ref{fig:missed_deadline_branch_75}
show that, when the accuracy requirement is very low, the conventional DNN may outperform the calibrated one. This is because, as seen in Figure~\ref{fig:accuracy_total}, the total accuracy for $p_{\text{tar}} = 0.75$ is larger than $p_{\text{tar}}$ for both conventional and calibrated DNNs. Therefore, calibration is not needed to ensure the reliability target, and it may be preferable to offload more often in order to meet the latency deadline by using a conventionally trained DNN. Note, however, that, both approaches yield the same missed deadline probability for a sufficiently large latency target $t_{\text{tar}}$. 

Figures~\ref{fig:missed_deadline_branch_825}~and~\ref{fig:missed_deadline_branch_85} consider the larger accuracy targets $p_{\text{tar}}=0.825$ and $p_{\text{tar}}=0.85$, respectively. In both figures, as long as the latency target is sufficiently large so as to enable a missed deadline probability lower than $0.2$, calibration significantly outperforms the conventional approach. The gains are particularly significant for the larger accuracy target of $p_{\text{tar}}=0.85$, under which the conventional approach is unable to obtain missed deadline probabilities lower than around $0.7$, while a calibrated DNN can reach $0.1$ for sufficiently large latency targets. 

Overall, the presented results on the missed deadline probability show the critical importance of calibrating a DNN even when the design is subject to sufficiently strict accuracy and latency requirements. 

\subsection{Impact of a Second Side Branch}
\label{subsec:secondsideBranch}

The analysis has so far considered a DNN with a single side branch. We now study the performance for a network with two side branches at the device. We specifically insert a second side branch after the second ReLU layer of B-AlexNet, i.e., between the second and the third convolutional layers. 

\begin{figure}[ht!]
\centering
\subfigure[$p_{\text{tar}}=0.825$.]{\label{fig:missed_deadline_branch_825_two}\includegraphics[width=0.75\linewidth]{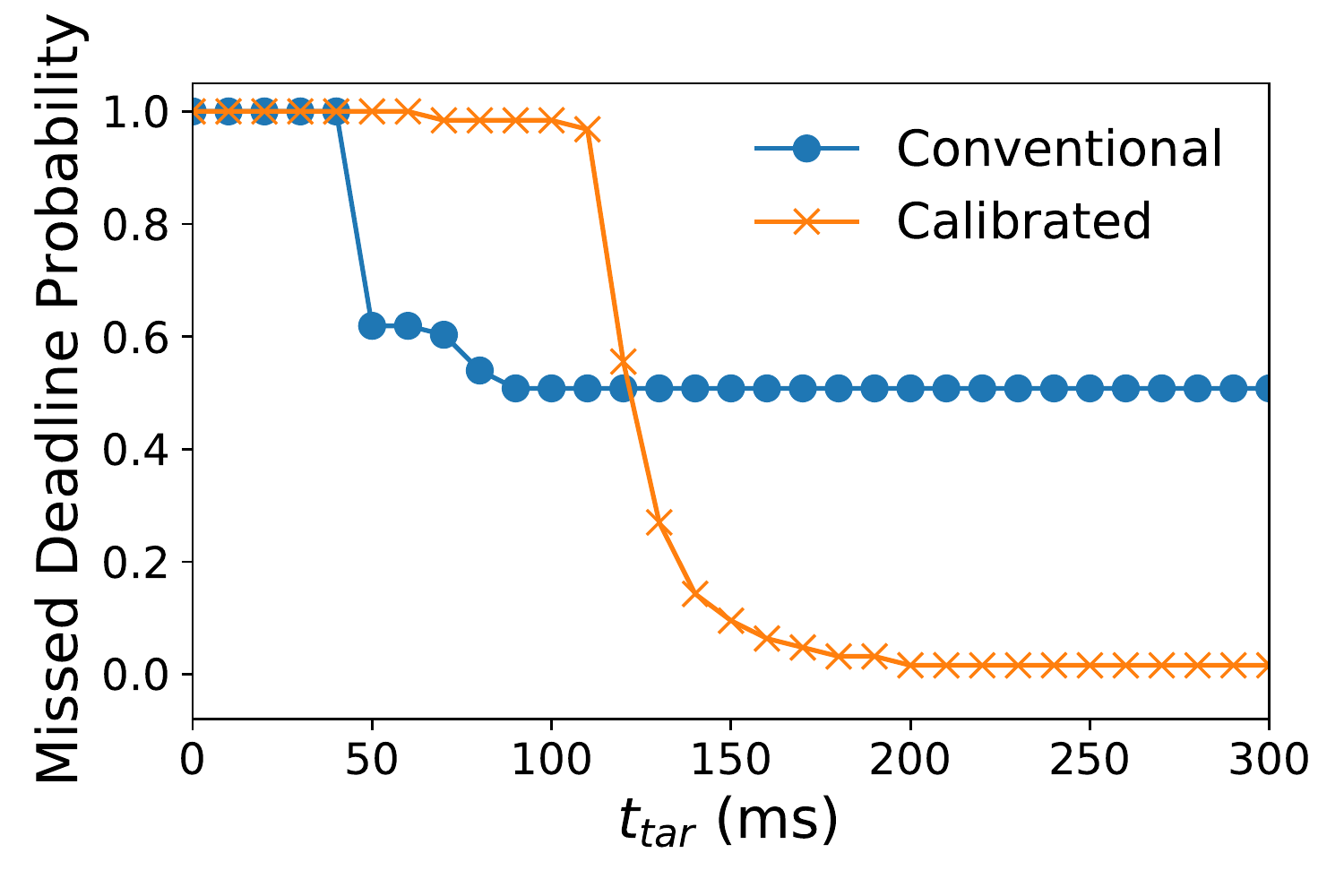}}
\subfigure[$p_{\text{tar}}=0.85$.]{\label{fig:missed_deadline_branch_85_two}\includegraphics[width=0.75\linewidth]{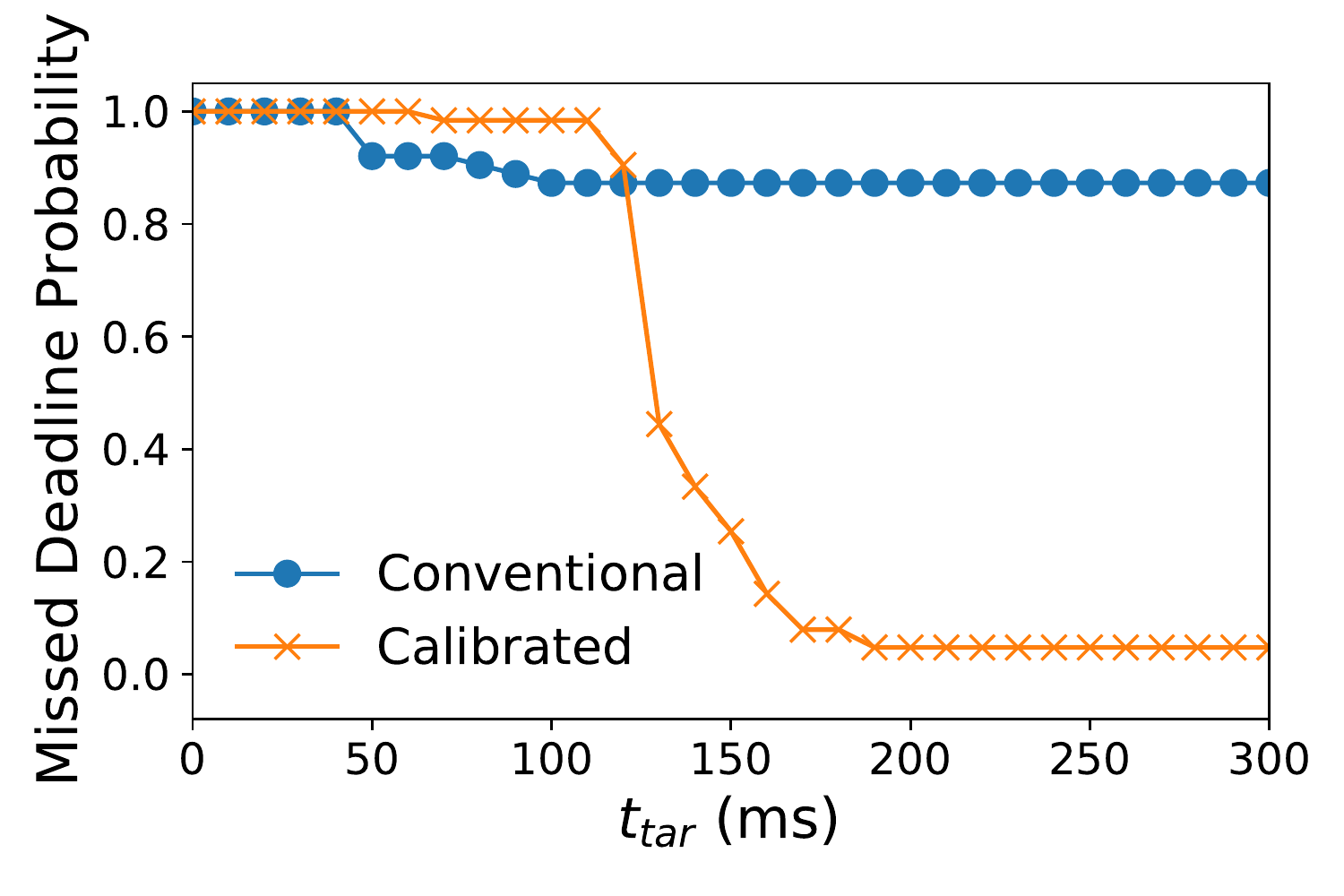}}
\caption{Missed deadline probability for a conventional and a calibrated two-branch DNN.}
\label{fig:missed_deadline_calibration_two}
\end{figure}

We start by considering end-to-end performance in terms of missed deadline probability by studying the same conditions and methodology of the last experiment presented in Section~\ref{subsec:missed_deadline}. We omit the results for $p_{\text{tar}} = 0.75$ since, as we discussed, this target is too low to make calibration effective. Apart from confirming the benefits of calibration, Figure~\ref{fig:missed_deadline_calibration_two}, when compared to Figure~\ref{fig:missed_deadline_calibration}, demonstrates that a conventional two-branch DNN obtains a larger missed deadline probability than its one-branch counterpart for sufficiently large $t_\text{tar}$. This is instead not the case for a calibrated DNN. This comparison indicates that, in the presence of more branches, a miscalibrated DNN has more opportunities to make unreliable offloading decisions, increasing the missed deadline probability.

\begin{figure}
\centering
\includegraphics[width=0.75\linewidth]{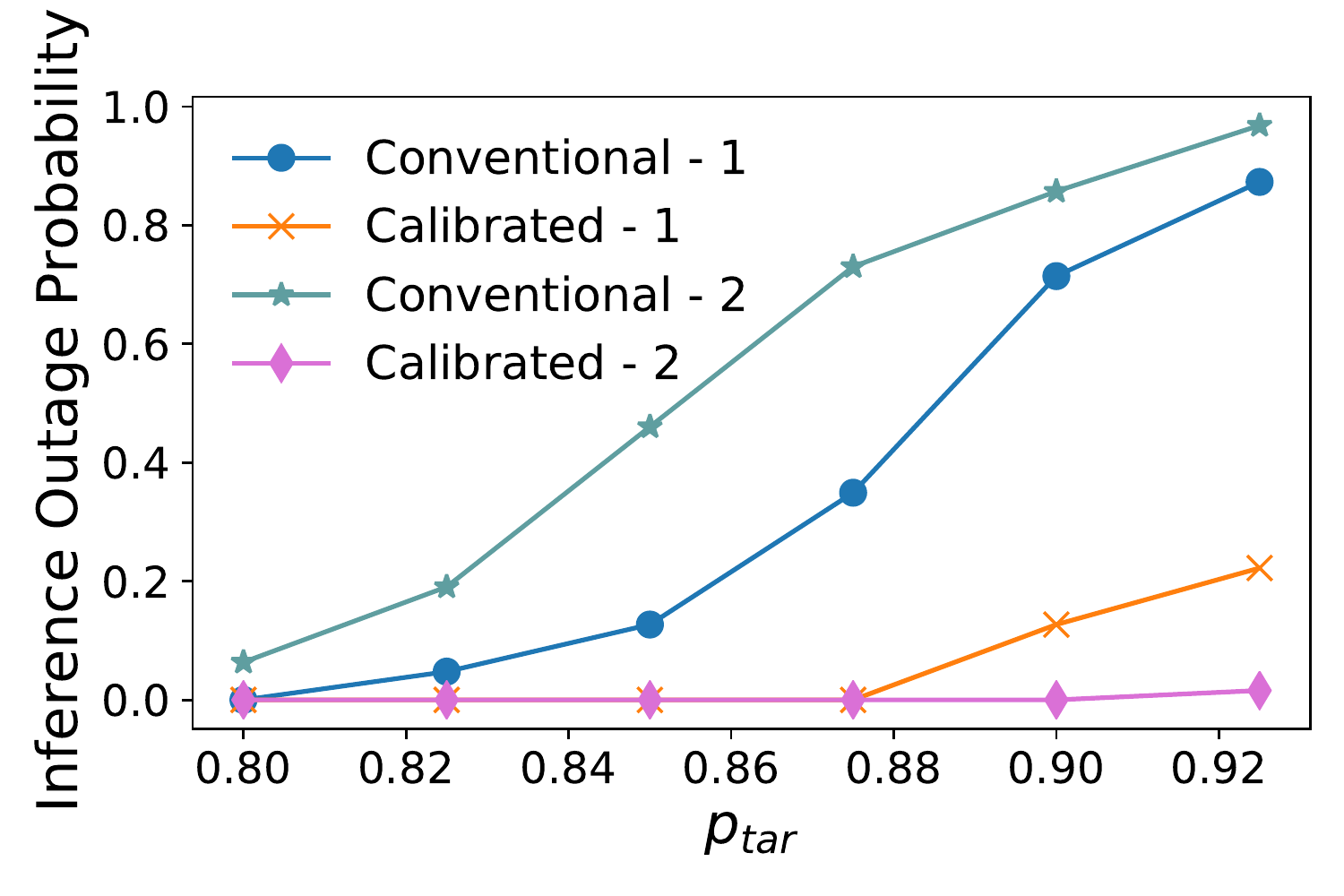}
\caption{Inference outage probability on the device, using conventional and calibrated one-branch and two-branch DNNs.}
\label{fig:outage_twobranch}
\end{figure}

To analyze the impact of more branches on device-level performance, Figure~\ref{fig:outage_twobranch} shows the inference outage probability as a function of the threshold $p_{\text{tar}}$ for conventional and calibrated DNNs with one and two branches. We obtain these results following the methodology of Section~\ref{subsec:probability_outage}. Figure~\ref{fig:outage_twobranch} demonstrates that a conventional DNN with two side branches has a larger inference outage probability than a one-branch conventional DNN. In contrast, for a calibrated DNN, the addition of a second side branch is beneficial and leads to a very low inference outage probability. This low outage probability, in turn, leads to the missed deadline probability levels shown in Figure~\ref{fig:missed_deadline_calibration_two}.   

\section{Conclusions}
\label{sec:conclusion}

An effective way to enable mobile offloading of DNN inference from devices to a cloud processor is model partitioning, which can be made adaptive via the introduction of early-exit side branches. In order to enable offloading decisions, a DNN needs to provide reliable estimates of confidence. These estimates are useful to decide whether to continue processing at the cloud or to return the result evaluated at the device. This work has investigated the impact of miscalibration on DNNs with early exits. We have provided extensive empirical evidence regarding miscalibration's role in providing unreliable offloading decisions. We have also demonstrated that the problem can be solved by applying a simple calibration method such as Temperature Scaling on a side branch. 

From a methodological standpoint, this paper introduced inference outage probability as a metric of reliability for inference on the device and the missed deadline probability to quantify the end-to-end performance of early-exit DNNs accounting for both accuracy and overall inference time. We have shown that calibration plays an essential role in meeting sufficiently tight application requirements, especially in the presence of multiple side branches.

Future work may evaluate how inference outage and missed deadlines affect the quality of experience for different DNN applications. Also, it would be of interest to consider other application requirements, such as energy consumption.

\section*{Acknowledgements}
This study was financed in part by the Coordenação de Aperfeiçoamento de Pessoal de Nível Superior - Brasil (CAPES) - Finance Code 001. It was also supported by CNPq, FAPERJ Grants E-26/203.211/2017, E-26/203.105/2019, and E-26/211.144/2019, and FAPESP Grant 15/24494-8. The work of O. Simeone has  received  funding  from  the  European  Research  Council  (ERC)  under  the  European  Union’s Horizon 2020 Research and Innovation Programme (Grant Agreement No. 725731).

\bibliographystyle{IEEEtran}
\bibliography{header,bibFile}

\end{document}